%% file: paper.tex
\documentclass[journal]{IEEEtran}
\usepackage{amsmath,amsfonts}
\usepackage{algorithmic}
\usepackage{algorithm}
\usepackage{array}
\usepackage[caption=false,font=normalsize,labelfont=sf,textfont=sf]{subfig}
\usepackage{textcomp}
\usepackage{stfloats}
\usepackage{url}
\usepackage{verbatim}
\usepackage{graphicx}
\usepackage{cite}
\usepackage{bbding}
\usepackage[inkscapelatex=false]{svg}
\usepackage{color}
\usepackage{xcolor}
\newcommand{\red}[1]{\textcolor{black}{#1}}
\newcommand{\zc}[1]{\textcolor{black}{#1}}

\begin{document}
\title{\red{Data-driven Methods} Applied to Soft Robot Modeling and Control: A Review}
\author{Zixi Chen, Federico Renda,~\IEEEmembership{Member,~IEEE,} Alexia Le Gall, Lorenzo Mocellin, Matteo Bernabei, Théo Dangel, Gastone Ciuti,~\IEEEmembership{Senior member,~IEEE,} Matteo Cianchetti,~\IEEEmembership{Member,~IEEE,} and Cesare Stefanini,~\IEEEmembership{Member,~IEEE}
\thanks{This work was supported in part by the European Union by the Next Generation EU project ECS00000017 ‘Ecosistema dell’Innovazione’ Tuscany Health Ecosystem (THE, PNRR, Spoke 4: Spoke 9: Robotics and Automation for Health.) and in part by the Khalifa University of Science and Technology under Grant CIRA-2020-074 and Grant RC1-2018-KUCARS. \textit{(Corresponding author: Zixi Chen.)} This work has been submitted to the IEEE for possible publication. Copyright may be transferred without notice, after which this version may no longer be accessible. }
\thanks{Zixi Chen, Alexia Le Gall, Lorenzo Mocellin, Matteo Bernabei, Théo Dangel, Gastone Ciuti, Matteo Cianchetti, and Cesare Stefanini are with the Biorobotics Institute and the Department of Excellence in Robotics and AI, Scuola Superiore Sant’Anna, 56127 Pisa, Italy (e-mail: zixi.chen@santannapisa.it; alexiamarie.legall@santannapisa.it; lorenzo.mocellin@santannapisa.it; matteo.bernabei@santannapisa.it; theo.dangel@santannapisa.it; gastone.ciuti@santannapisa.it; matteo.cianchetti@santannapisa.it; cesare.stefanini@santannapisa.it).}
\thanks{Federico Renda is with the Khalifa University Center for Autonomous Robotic Systems, Khalifa University, Abu Dhabi 127788, United Arab Emirates (e-mail: federico.renda@ku.ac.ae).}
}
\maketitle

\begin{abstract}
Soft robots show compliance and have infinite degrees of freedom. Thanks to these properties, such robots can be leveraged for surgery, rehabilitation, biomimetics, unstructured environment exploring, and industrial grippers. In this case, they attract scholars from a variety of areas.
However, nonlinearity and hysteresis effects also bring a burden to robot modeling. Moreover, following their flexibility and adaptation, soft robot control is more challenging than rigid robot control.  
In order to model and control soft robots, a large number of data-driven methods are utilized in pairs or separately.
\zc{This review first briefly introduces two foundations for data-driven approaches, which are physical models and the Jacobian matrix, then summarizes three kinds of data-driven approaches, which are statistical method, neural network, and reinforcement learning.} This review compares the modeling and controller features, e.g., model dynamics, data requirement, and target task, within and among these categories. 
Finally, we summarize the features of each method. A discussion about the advantages and limitations of the existing modeling and control approaches is presented, and we forecast the future of data-driven approaches in soft robots.
A website (\url{https://sites.google.com/view/23zcb}) is built for this review and will be updated frequently.
\end{abstract}

\def\abstractname{Note to Practitioners}
\begin{abstract}
This work is motivated by the need for a review introducing soft robot modeling and control methods in parallel.
Modeling and control play significant roles in robot research, and they are challenging especially for soft robots. The nonlinear and complex deformation of such robots necessitates specific modeling and control approaches. 
We introduce the state-of-the-art data-driven methods and survey \zc{three} approaches widely utilized. This review also compares the performance of these methods, considering some important features like data amount requirement, control frequency, and target task. 
The features of each approach are summarized, and we discuss the possible future of this area.
\end{abstract}


\section{Introduction}
\label{sec1}
\IEEEPARstart{S}{oft} robots have been developed for a large number of applications.
Owing to their infinite degrees of freedom (DOFs) and flexibility, soft robots are leveraged in robot assistant surgery, especially minimally invasive surgery \cite{13MC}.
Compared with their rigid counterparts, soft robots are relatively safe due to their softness and have significant advantages in assisting elderly and disabled people with daily tasks \cite{18YA} and cooperating with humans\cite{14JQ}.
Moreover, they are used as hand recovery devices\cite{22ZT} and wearable devices\cite{18JT} for various medical purposes like rehabilitation and human motion monitoring.
Animals are composed of soft tissues, and researchers in the bioinspired area apply soft materials like silicone to build soft robots and mimic the behaviors of living beings, such as octopus\cite{13CL}, elephant\cite{21SJ}, and earthworm\cite{04AM}. These robots produce specific motions and manipulations by imitating the structures and behaviors of soft animals.
With the help of such soft biomimetic robots, some exploring tasks in various environments like underwater\cite{18RK} and walls\cite{23JS} can be achieved.
Soft robot hands are adaptative to objects and commonly applied for grasping tasks involving fragile\cite{16KG} and diverse\cite{22CF} objects.
To summarize, unique properties like infinite DOFs, compliance, and safety for humans lead to the high potential of soft robots. Hence, soft robot study is a research area highly attractive to robotics scholars.

\red{Followed by the aforementioned advantages and applications, the most considerable drawback of soft robots is the challenge of modeling and control. Deformable materials lead to the nonlinear and delayed responses of soft robots. Moreover, infinite DOFs make it complicated to build accurate models for soft robots. The physical properties of soft robots will also vary due to the aging of soft materials. Owing to the above characteristics, the modeling of soft robots is more complex than that of their rigid counterparts, which also leads to challenging control tasks\cite{20KC, 22JW}. Therefore, modeling and control play essential roles in soft robot research.}

\begin{figure*}[!ht]
\centering
\includegraphics[width=6.0in]{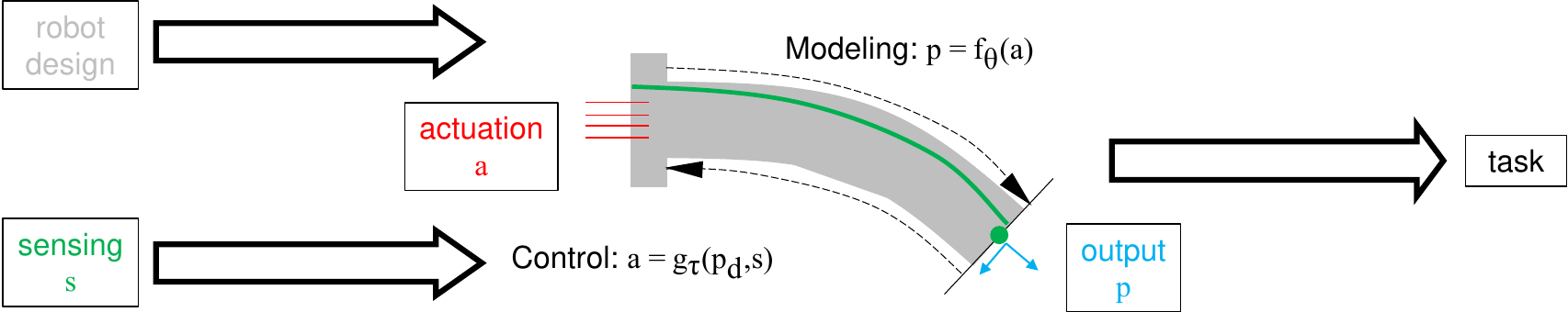}
\caption{\red{Diagram of soft robot modeling and control processes.} Robot design (grey) provides a specific soft robot structure and actuation pattern. The sensing system (green) obtains robot information via the sensors of such a robot. \red{Modeling $f_\theta$ is a function that predicts the robot state $p$ (blue), as end position and robot pose, according to the actuation variables $a$ (red.) Control $g_\tau$ aims to decide actuation variables $a$ with the desired state $p_d$ and sensing input$s$. Finally, such a robot system can achieve a variety of tasks. $\theta$ and $\tau$ represent the parameters in the data-driven methods of modeling and control, respectively.}}
\label{fig1}
\end{figure*}

\input{tables/table_sum}

The working process of soft robots is summarized in Fig. \ref{fig1}.
\red{Various kinds of robot designs are used in soft robots, like 1 DOF soft fingers\cite{20ZT}, 3 DOF continuum robots\cite{17KL}, parallel robots\cite{13CD}, concentric tube robots\cite{10PD} and so on.
To actuate a soft robot, the researchers apply actuation approaches like fluid-driven methods\cite{18TT}, cable-driven method\cite{16MY}, electroactive polymers\cite{16HW}, shape memory alloy (SMA)\cite{19CC}, etc.}
Also, there are multiple categories of sensors, e.g., optical markers\cite{19TT}, EM trackers\cite{22XHb}, flex sensors\cite{20JJ} and Fiber Bragg Grating (FBG)\cite{22XH,22ZD}.
Some reviews detailedly introduce soft robots with mechanical aspects\cite{17CL,22KL}.
\red{Based on these hardware implementations, soft robot modeling can be seen as a function that takes actuation $a$ as the input and robot state $p$, such as end position, orientation, and shape, as the output. Meanwhile, soft robot control is the inverse process of modeling, which takes the desired robot state $p_d$ and the sensing signal $s$ as the input and actuation $a$ as the output.
The input and output choices may change considering the control strategy. For example, the open control strategy in \cite{20JB} only requires the target positions $p_d$ for the trajectory following, but the close loop control approach in \cite{19TT} utilizes both the target position $p_d$ and the previous end positions from the sensing system $s$.
Each mapping requires a model, $f$ or $g$, and the parameter $\theta$ or $\tau$, which are partly influenced by robot design and sensing system.}
Finally, such control methods are leveraged for simple tasks, like target reach\cite{23KI} and trajectory following\cite{22DW}, and challenging tasks, like interaction adaptation\cite{14JQ} and navigation\cite{14AT}.

\red{Due to the variety of structures and the complexity of behaviors, it is challenging to propose accurate physical models and corresponding control strategies for every soft robot. However, data-driven methods have shown considerable benefits. 
Data-driven approaches summarize the features of data collected from robot motions without the necessity of robot design knowledge. Furthermore, compared with physical approaches based on simplifications and hypotheses, data from real robots can illustrate the real features of soft robots.
Thanks to these advantages, some data-driven approaches can be proposed for robot modeling and control with optimization\cite{14MY} or learning\cite{18TT}. 
Data-driven approaches can be applied for various kinds of modeling and control strategies, like kinematics modeling\cite{14AM}, dynamic modeling\cite{18MG}, open loop control\cite{19SS}, and close loop control\cite{22AC}.
Moreover, they can also be utilized for observer \cite{18ML} which includes modeling as prediction and sensing characterization\cite{23XW} which can be applied in control.}

\begin{figure*}[!ht]
  \subfloat[]{\includegraphics[width=0.5\textwidth]{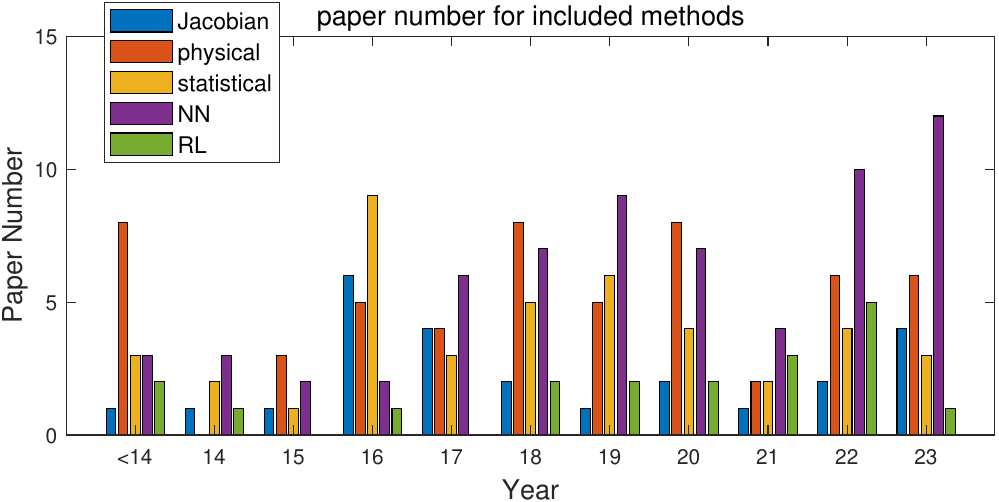}}
  \subfloat[]{\includegraphics[width=0.5\textwidth]{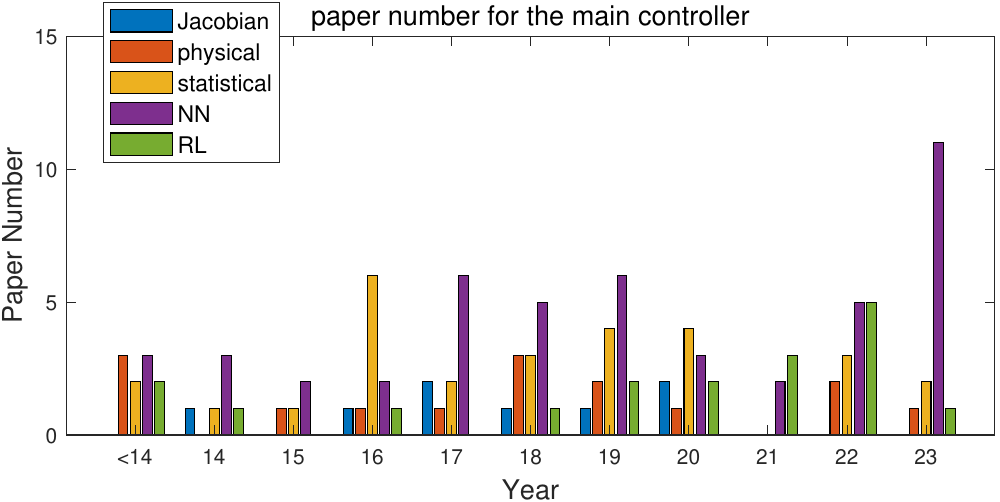}}
\caption{Paper number for (a) included methods and (b) \red{the main method}}
\label{fig2}
\end{figure*}

\begin{figure*}[!ht]
  \subfloat[]{\includegraphics[width=0.5\textwidth]{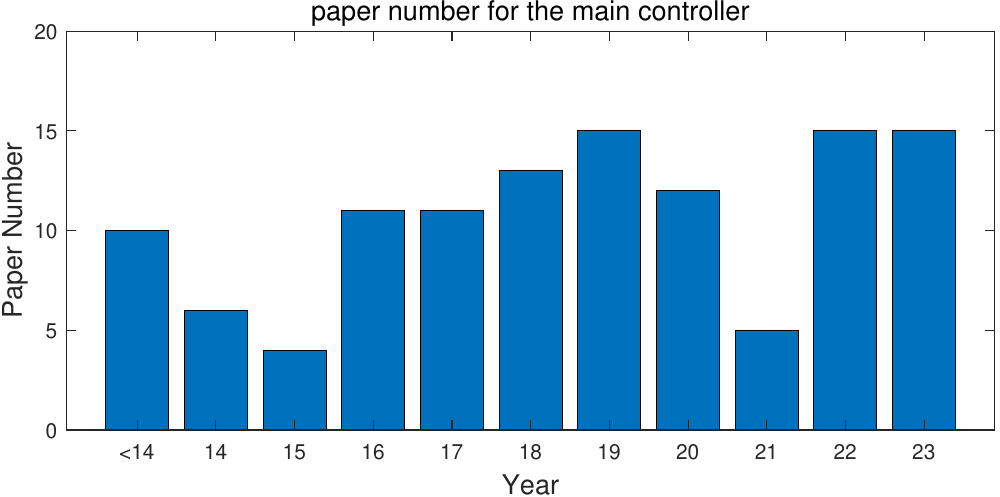}}
  \subfloat[]{\includegraphics[width=0.5\textwidth]{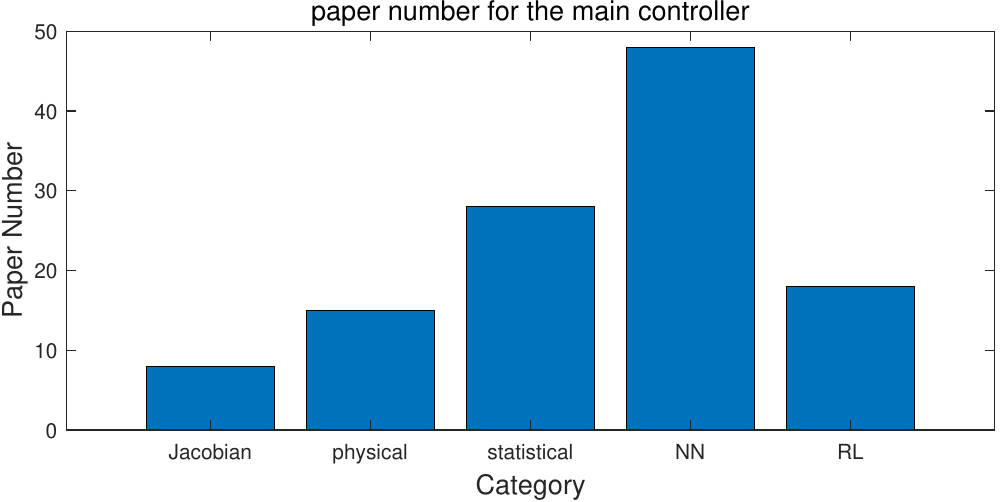}}
\caption{Paper number for \red{the main method} according to (a) years and (b) categories}
\label{fig3}
\end{figure*}

This review aims to present and compare \red{data-driven methods} applied to soft robot modeling and control in parallel. 
Although some reviews are related to soft robot modeling and control\cite{18TTc,21CS,22JW,23CL}, they do not introduce modeling and control \red{simultaneously}.
\zc{In this review, first we briefly introduce foundations for data-driven approaches, which are physical models and the Jacobian matrix. Physical models provide simulation environments and insight into soft robots, while the Jacobian matrix infers the data relationship for data-driven approaches.}
\red{Then, we classify data-driven methods applied to soft robot modeling and control into three categories: statistical method, neural network, and reinforcement learning.}
We introduce the corresponding modeling and control approaches for each kind of \red{data-driven approach} parallelly and compare them within and among these categories. Finally, conclusions on \red{data-driven approaches} applied to soft robot modeling and control are proposed, and we discuss the challenges and emerging directions of this area. 
Fig. \ref{fig2} shows the paper number of different \red{methods} cited in this review according to the publishment time. In Fig. \ref{fig2}-(a), one paper may fall into multiple categories if it involves several methods, but one paper only falls into one category considering the main method in this paper in Fig. \ref{fig2}-(b). Fig. \ref{fig3}-(a) and Fig. \ref{fig3}-(b) summarize the number of papers according to the publishment time and \red{method} category, respectively. The summary of \red{data-driven method} categories is shown in Table \ref{table_sum}.

\section{\zc{Foundations for Data-driven Approaches}}
\label{sec2}
\zc{In this section, we introduce two significant foundations for data-driven methods, which are physical models and the Jacobian matrix. Both of them are very popular and effective in rigid robot modeling and control. In this case, the researchers also try to employ them in soft robots. 
They stimulate the development of data-driven approaches by providing simulators or inferring data relationships.}

\subsection{\zc{Physical Models}}
\label{sec2.1}
\begin{figure}[!ht]
\centering
\includegraphics[width=3.0in]{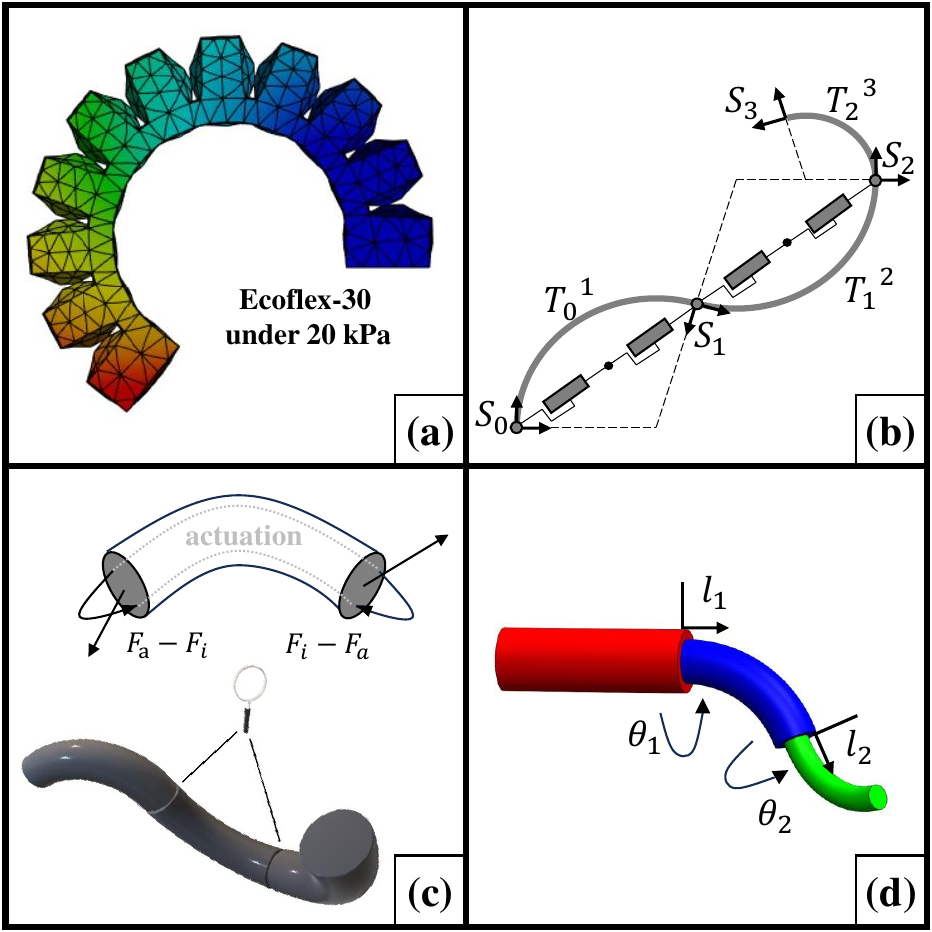}
\caption{Diagrams of (a) FEM, (b) PCC with pseudo-rigid model, (c) Cosserat rod model, and (d) concentric tube model. \red{Element motion and deformation in (a) represent the soft robot deformation. The grey soft robot in (b) is represented by a series of constant curves, and the augmented rigid robots model the soft robot dynamically. The Cosserat rod model in (c) includes forces and moments inherently. The concentric tube model in (d) represents the robot motion with the configurations $l_1,l_2,\theta_1$ and $\theta_2$.}}
\label{fig4}
\end{figure}

\zc{Physical models are significant in soft robotics because they can illustrate the nature of soft robot motion and deformation. Also, they are fundamental to data-driven methods by providing simulation environments and data.} Discretization methods like FEM have been applied for robot simulation. The Cosserat model and its discretization method, like geometric variable-strain (GVS), provide simulators specifically for soft robots. Piecewise constant curvature (PCC) can be seen as a special case of GVS and is widely applied to the soft robot simulation based on the assumption of deformation shape. Some specific models, like the pseudo-rigid model and the concentric tube model, are proposed for specific soft robots.
Some typical physical models are shown in Fig. \ref{fig4}. \zc{In robot control, methods considering physical models or parameters are model-based, while approaches only employing data and not using physical relationships inherent in robots can be seen as model-free\cite{21BZ}.}

\red{FEM is a popular discretization simulation method in robotics. This method can be applied to various soft robots, for example, soft foam robot hand\cite{18CS}, soft pneumatic actuator\cite{17GR}, and PneuNet bending actuator\cite{18MW}.
This model can provide high-accuracy simulation with the requirement of material parameters such as mass density, Young’s modulus, and Poisson’s ratio. 
Generally, FEM works as an offline modeling method, such as producing a dataset\cite{17GR} for NN learning and building an exploring environment for RL \cite{18CS}.
FEM is applied in the simulator SOFA\cite{12FF} for deformation and interaction simulation.
Moreover, a series of works \cite{13CD,15FL,19MT} focus on FEM applications for real-time control. In order to cope with this highly complicated model, methods like local linearization\cite{13CD} and reduced-order control model\cite{19MT} are employed for the trade-off between the model accuracy and control frequency.}

Besides the discretization of 3D small elements, the discretization of 1D Cosserat rod is also applied. Cosserat rod models soft continuum robots with a series of rigid cross-sections, and includes bend, twist, stretch, and shear. The simulator PyElastica\cite{18MGb} applies a discrete geometric form of the Cosserat rod model for soft robot simulation. The most general strain-based discretization method of the Cosserat rod is GVS\cite{20FR}, which discretizes the continuous Cosserat rod model into a finite set of strain basis functions. SoRoSim\cite{22AM} integrates GVS for soft, rigid, and hybrid robotic system simulation. PCC is a special GVS, which only includes bending and simulates a continuum robot with several curves in some bending angles\cite{17HJ}. Original PCC only shows the geometrical information and dynamic PCC models include spatial dynamic effects\cite{16IG} and external interactions\cite{21EM}. \zc{Of note, PCC is a simplification of the nonlinear Euler-Bernoulli beam, which can be applied for soft robot modeling under external interaction without the constant curvature assumption\cite{23MG}.}

\red{There are also some physical models for specific kinds of soft robots. Considering concentric tube robots, the authors of \cite{10PD} leverage rotations and translations of tubes as actuation variables for the concentric tube robot model, as shown in Fig. \ref{fig4}-(d). The control tasks are achieved by finding the inverse kinematic solutions. Neural networks are utilized for both modeling and control in \cite{18RG} based on this model. The concentric tube model has also been used for RL control in \cite{23KI} to provide action variables. Pseudo-rigid models can approximate soft robots with rigid counterparts\cite{18CSb} as shown in Fig. \ref{fig4}-(b), and the sophisticated motions of soft robots are simulated with the help of pseudo springs\cite{19ZT} and dampers\cite{17ZW}. }

In conclusion, the research of physical models deepens the understanding of soft robot nature and \red{produce various soft robot simulators for data-driven approaches}.
A comparison of some typical papers applying physical models is shown in Table \ref{table_p}. \red{This table includes the simulator SOFA based on FEM\cite{12FF}, the simulator PyElastica based on Cosserat rod \cite{18MGb}, the simulator SoRoSim based on GVS\cite{22AM}, and RL works using the specific model concentric tube model\cite{23KI}.}
\input{tables/table_p}

\subsection{\zc{Jacobian matrix}}
\label{sec2.2}

The Jacobian matrix can infer the relationship between the actuation and position velocities \zc{by model linearization}. 
Thanks to its concision and effectiveness, \zc{the Jacobian matrix is widely applied in rigid robots, whose} explicit physical models are easy to propose. Also, it is possible to linearize such models.
However, due to the difficulties of building accurate and explicit models for soft robots, optimization is utilized for Jacobian matrix estimation instead of directly linearizing explicit models. 
The included robot model is just a general model like $p=f_{\theta}(a)$ instead of an explicit and physical one, and \zc{it is only employed for illustration instead of calculation\cite{14MY}.}

\begin{figure}[!ht]
\centering
\includegraphics[width=3.0in]{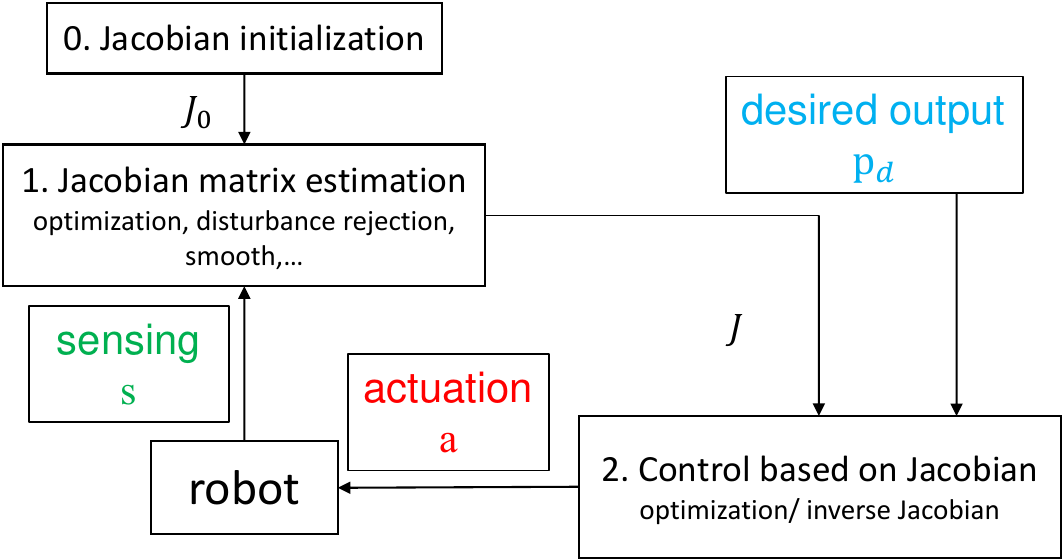}
\caption{\zc{diagram of the Jacobian matrix estimation and control}. Based on the Jacobian matrix initialization, the Jacobian matrix is updated based on the sensing $s$. Then, the estimated Jacobian matrix is applied for control.}
\label{fig5}
\end{figure}

The \zc{Jacobian matrix estimation} and control process is summarized in Fig. \ref{fig5}. \zc{The first work applying the Jacobian matrix in the main approach is \cite{14MY}.} For Jacobian matrix initialization, each actuator should be actuated with a small incremental amount in order to estimate the initial Jacobian matrix $J_0$\cite{14MY,18XW,20XW,20YW}. Then, the Jacobian matrix is updated according to the actuation and end position change in the last step. The matrix update strategy can be shown as
\begin{equation}
\label{eq3_1}
\begin{split}
\min_{\hat{J}^{k+1}}\ &{\Vert\triangle\hat{J} \Vert}_2\\
s.t.\ &\triangle x^k=\hat{J}^{k+1}W\triangle y^k\\
&\hat{J}^{k+1} = \hat{J}^{k}+\triangle\hat{J}
\end{split}
\end{equation}
\red{where $\triangle x^k$ and $\triangle y^k$ are the end effector displacement and actuation change at step $k$, respectively.} $J=[J_1\ J_2\ \dots\ J_n]$ is the Jacobian matrix and $n$ is the dimension of the actuation variable $y$. $W=diag({\Vert J_1\Vert}_2,{\Vert J_2\Vert}_2,\dots,{\Vert J_n\Vert}_2)$ is a weighting matrix and $\hat{J}=JW^{-1}$ is the unit Jaocbian matrix. $\hat{J}^{k}$ and $\hat{J}^{k+1}$ are the unit Jacobian matrix at step $k$ and $k+1$.

\red{The actuation variables are derived as the cost function in an optimization problem considering the constraints of the Jacobian matrix and target end position for control.}
The control strategy for step $k+1$ can be shown as
\begin{equation}
\label{eq3_2}
\begin{split}
\min_{\triangle{y}^{k+1}}\ &\red{{\Vert y^k+\triangle y^{k+1} \Vert}_2}\\
s.t.\ &\triangle x_d=\hat{J}^{k+1}W\triangle y^{k+1}\\
&\red{y^k+\triangle y^{k+1}\geq y_{min}}\\
&\red{y^k+\triangle y^{k+1}\leq y_{max}}
\end{split}
\end{equation}
where $\triangle x_d$ is the desired displacement for the end effector. The actuation value $y^k+\triangle y^{k+1}$ is constrained between the minimal and maximal actuation value $y_{min}$ and $y_{max}$ according to the robot structure. Except for \red{the optimization control method in Eq. \ref{eq3_2}}, the inverse Jacobian matrix is also utilized for control \cite{18ML}.


Considering \zc{nonlinearity and hysteresis of soft robots}, the Jacobian matrix in the last step may not be accurate and suitable for this step. It is challenging for the original Jacobian \zc{estimation method} to face complex tasks due to its oversimplification. Therefore, some researchers try to adapt this method for various applications. For example, the authors of \cite{20YW} also emphasize that there may be a significant deviation between the estimated and real Jacobian matrix, and they alleviate this issue by smoothing the activation values. \zc{Force-displacement model is included for force control in \cite{16MY}.} When end effector and actuation displacement between only one step is involved in \cite{14MY}, the displacement among multiple steps is utilized in \cite{17MY}.
Also, the Jacobian matrix is adjusted in \cite{19MV} by multiplying a rotating matrix considering the difference between the intended and measured motions. \zc{The fusion of sensing information from a camera and FBG provides accurate positions in \cite{20XW} for stable and precise Jacobian matrix estimation. }

\zc{Although so many adaptations have been proposed, the Jacobian matrix is still an initial solution for many labs and motivates applying the other approaches, especially data-driven ones.}
\zc{A neural network is applied for control in \cite{17TTb}, which is based on the Jacobian matrix estimation process.}
A honeycomb pneumatic networks arm is controlled by the inverse Jacobian matrix in \cite{17YJ}, and the same soft robot is included in \cite{21HJ} for challenging manipulation tasks. The latter work employs the Jacobian \zc{matrix in} a low-level behavior controller and compares it with RL.


In summary, the Jacobian matrix is a concise and simple \red{method} for soft robots.
\red{The matrix is updated online with a high frequency in most cases, which can be achieved due to the simple structure. Meanwhile, the oversimplified linearization necessitates online updating and high control frequency. 
A comparison of some typical papers using the Jacobian matrix is shown in Table \ref{table_j}. This table contains the first paper employing the Jacobian matrix for soft robot control\cite{14MY}. Force control is achieved in \cite{16MY} considering a force model in the control optimization problem. The authors of the last work \cite{21HJ} take inspiration \zc{from the Jacobian matrix in\cite{17YJ}} and implement RL based on the same robot.}
\input{tables/table_j}

\section{Statistical \red{Method}}
\label{sec4}
\red{Statistical methods are utilized to build the mapping functions between different variable spaces with only data, and physical relationships among these spaces are unnecessary.
For example, the authors of \cite{19GF} infer the relationship between the actuation input and image feedback for kinematic control, and the authors of \cite{16HW} plan the actuation variables based on the temporal values.}
\red{There are many regression approaches utilized in soft robots, like linear regression\cite{19SSb}, local weight regression (LWR)\cite{16WXb}, support vector regression (SVR)\cite{20IL}, Gaussian process regression (GPR)\cite{20JJ}, and local weight projection regression (LWPR)\cite{17KL}. Other than the regression methods, the Gaussian mixture model (GMM)\cite{19BY} summarizes collected data with a joint data distribution, and the extended Kalman filter (EKF)\cite{18ML} estimates robot states as an observer. }

\subsection{Regression Method}
\label{sec4.1}
\red{Regression methods with different models are employed in soft robot modeling and control. These methods aim to fit the training data with a specific model, like a linear function or a Gaussian process, by optimizing the parameters and decreasing the loss. Then, the trained regression model can take some observation samples as model input and predict the corresponding values.
For instance, linear regression, a simple regression approach, is applied in \cite{19SSb} to map FBG signals into soft robot end position for sensing. A linear function is included for fitting, and the parameter matrix is optimized. Similarly, LWR employs the linear function but considers the distance between the collected data and the observation samples for fitting. In \cite{16WXb}, data from human demonstrations is used for fitting. The temporal value is taken as input to decide action for control.}

SVR is utilized in \cite{20IL} for the forward kinematic modeling and in \cite{17AM} for the close loop position controller. It has been proven in \cite{17AM} that SVR gets better approximation accuracy than NN on a simple function, but this model requires more convergence time than NN on a large amount of data (15625 samples) in \cite{20IL}, which may be caused by the different optimization strategies or mature NN optimization software. \red{The SVR modeling and control algorithm in \cite{17AM} can be summarized as 
\begin{equation}
\label{eq4_1}
\begin{split}
\min_{W, B}\ & \frac{1}{2}\sum^M_{j=1}{\Vert w_j\Vert}^2+C\sum^n_{i=1} L(\mu_i),\\
s.t.\ &\mu_i=\Vert y_i-(\phi(x_i)W+B)\Vert\\
&W={[w_1,w_2,\dots,w_M]}\\
\end{split}
\end{equation}
where $d,M$ are the dimensions of the mapping input $x_i$ and output $y_i$, and $n$ is the size of the learning dataset. The function $\phi(x_i)W+B$ aims to estimate the mapping output with a nonlinear transformation $\phi(\cdot):R^{1\times d}\rightarrow R^{1\times h}$ and the optimized matrices $W\in h\times M$ and $B\in1\times M$. $L(\cdot)$ is a loss function and the extended Vapnik $\epsilon$-insensitive loss function based on L2-norm is applied in \cite{17AM}. $C$ is the tradeoff parameter that adjusts the estimation errors and regularization.
For kinematics modeling, SVR takes the actuation as input $x$ and estimates $y$ as the robot state. For control, which can be seen as the inverse process of modeling, SVR receives the desired displacement as input and decides the actuation.}

GPR employs the Gaussian process, a probability distribution over functions, for \red{fitting}. 
For modeling, GPR predicts robot states, such as position or shape parameters, based on the training dataset and current actuation variables. \zc{For example, GPR is applied in \cite{20JJ} to predict the actuator curvature. 
Given $N$ training inputs $X=[x_1,x_2,\dots,x_N]^T \in R^{N\times i}$ and $N$ curvatures as the training outputs $Y=[y_1,y_2,\dots,y_N]^T \in R^{N\times 1}$, where $i$ represents the dimension of each input $x$. The prediction process based on the test input $x_t$ can be shown as}
\begin{equation}
\label{eq4_5}
\begin{split}
&\zc{\mu(x_t)=k_*(X,x_t)^TK(X,X)^{-1}Y,}\\
&\zc{\Sigma(x_t)=k(x_t,x_t)-k_*(X,x_t)^TK(X,X)^{-1}k_*(X,x_t),}
\end{split}
\end{equation}
\zc{where $\mu(x_t)$ and $\Sigma(x_t)$ are the predictive mean and variance. $k(\cdot,\cdot)\in R$ is the kernel function applied in GPR, a squared-exponential kernel function in most cases, and $k_*(X,x_t)=[k(x_1, x_t), k(x_2, x_t),...,k(x_N, x_t )]^T \in R^{N\times1}$. $K(X,X)\in R^{N\times N}$ is a covariance matrix with entries $K_{ij}=k(x_i,x_j), i,j=1,2,\dots,N$.} \zc{The above prediction process supposes that the prior mean is zero, and one can preprocess the sample output $Y$ by zeroing the mean before fitting and prediction.} It should be noticed that the noise of the mapping is considered in \cite{19GF}, which is assumed as white Gaussian noise with zero mean and variance $\sigma_n^2$. In this case, the prediction can be derived as
\begin{equation}
\label{eq4_6}
\begin{split}
&\zc{\mu(x_t)=k_*(X,x_t)^T{(K(X,X)+\sigma_n^2I)}^{-1}Y,}\\
&\zc{\Sigma(x_t)=k(x_t,x_t)-k_*(X,x_t)^T{(K(X,X)+\sigma_n^2I)}^{-1}k_*(X,x_t).}\\
\end{split}
\end{equation}
With the forward model derived from GPR, the authors of \cite{22ZT} propose a control strategy by minimizing a cost function containing the predicted errors and actuation variables. GPR is also employed in close-loop kinematics control in \cite{19GF} by predicting desired actuation variables based on the robot state feedback. The authors of \cite{20ZT} aim to predict the difference of robot states instead of only the next states as the modeling part in optimal control.

Based on LWR, which utilizes each training data as a local model, LWPR projects training data into several linear models and weighs them with Gaussian kernels. The covariances of these kernels are decided by an incremental gradient descent based on stochastic leave-one-out cross validation criterion \cite{16GF,22ZTb}. To find the optimization of the parameters in linear models, which is a redundancy problem, a null-space behavior is defined as a guide. This user-defined behavior is applied in a cost function and attracts the actual behaviors. For example, the robot elongation is minimized in \cite{17KL} for a relatively straight shape, and the overall inflated chamber pressures are minimized during the control process in \cite{18JH}.

\subsection{Gaussian mixture model}
\label{sec4.2}
\red{GMM encodes collected data into a data distribution composed of multiple Gaussian components.} In the fitting step, GMM parameters are optimized to fit the collected data. During the prediction step, the input data works as the prior, and the posterior of GMM under some input data will be used as prediction output. 
Of note, once a GMM is built, each kind of data plays the same role in principle. In such a joint probability density function, every dimension can be applied as a prior and derive expectations on the remaining dimensions. In this case, GMM in \cite{19BY} produces both forward \red{kinematics} modeling and \red{position} control strategy with actuation variables and desired end positions as the prior, respectively. \red{The catheter kinematic GMM is represented by the joint probability density function:
\begin{equation}
\label{eq4_2}
\begin{split}
P[p[k+1],p[k],a],
\end{split}
\end{equation}
where $p[k]$ denotes the robot state at step $k$. The forward modeling process can be shown as
\begin{equation}
\label{eq4_3}
E[p[k+1] \vert p[k],a],
\end{equation}
which is the conditional mean of the model Eq. \ref{eq4_2} given the robot state $p[k]$ and actuation $a$. The control strategy is 
\begin{equation}
\label{eq4_4}
E[a \vert p_d[k+1], p[k]],
\end{equation}
which is the conditional mean of the model Eq. \ref{eq4_2} given the robot state $p[k]$ and desired state $p_d[k+1]$.}

In addition to modeling and control, such data encoding characteristic develops some planning solutions. For example, the authors of \cite{16HW} encode pose and temporal value into a GMM for navigating through narrow holes based on human demonstration. Moreover, other task parameters like the rotation matrix of the coordinate system are included \red{as the planning objects} in the GMM of \cite{16MM}. Making use of its encoding ability, GMM transfers demonstrations on rigid robots to the STIFF-FLOP continuum robot in \cite{14SC}.

\subsection{Extended Kalman filter}
\label{sec4.3}
\red{Considering one existing model and its prediction, the Kalman filter can be applied as an observer and corrects the predicted values with measurement. Due to the nonlinear responses from soft robots, most modeling process is nonlinear, and the extended Kalman filter (EKF), instead of the original linear Kalman filter, is widely applied in soft robots.} $P, Q, R$ represent the estimation covariance, process noise covariance, and measurement noise covariance, respectively. In prediction, the state at the $k+1$ step is predicted as $p_{k+1|k}$ based on the state $p_{k|k}$ and actuation $a_k$ at the $k$ step.
\begin{equation}
\label{eq4_7}
\begin{split}
p_{k+1|k}&=f(p_{k|k},a_k),\\
P_{k+1|k}&=A_kP_{k|k}A_k^T+Q_k,\\
\end{split}
\end{equation}
where $f(\cdot,\cdot)$ represents \red{the nonlinear} forward modeling, $A_k$ is the local linearization of $f(\cdot,\cdot)$. The correction process corrects the prediction state to $p_{k+1|k+1}$ considering measurement \red{output} $s_k$ from sensing.
\begin{equation}
\label{eq4_8}
\begin{split}
K_k&=P_{k+1|k}C_k^T{(C_kP_{k+1|k}C_k^T+R_k)}^{-1},\\
p_{k+1|k+1}&=p_{k+1|k}+K_k(s_k-g(p_{k+1|k})),\\
P_{k+1|k+1}&=(I-K_kC_k)P_{k+1|k},\\
\end{split}
\end{equation}
where $g(\cdot)$ represents the measurement process, $C_k$ is the local linearization of $g(\cdot)$. $K_k$ is the Kalman Gain and evaluates the reliability of measurement and prediction. EKF is commonly leveraged since it \red{adjusts the modeling process and provides an accurate robot state estimation.}
For example, the authors of \cite{16AA} map actuator variables and segment parameters into the robot pose to build the nonlinear forward modeling with the transformation matrices\red{, and the sensing signal from position sensors is applied for correction.}
The authors of \cite{16CJ} \red{aim to estimate robot poses and physical parameters and apply pose measurement for correction.}
NNs like wavelet networks \cite{19JL} are also employed as the forward model in EKF \red{for curvature angle estimation. In their following research, the authors of \cite{20JL} also estimate the external force as the unknown system input based on the state estimation from a similar EKF. Similarly, the unknown external forces are taken as the system state $p$ and accurately estimated in \cite{11DR}.} Furthermore, due to the modular structure of the snake robot in \cite{11ST}, the EKF is adapted by changing the dimension of state variables according to the advancing or retracting motion \red{for shape estimation}.

\subsection{Summary Statistical Method}
\label{sec4.4}
\red{Because statistical models only consider data relationships, such models have shown high potential to cope with various kinds of data. At first, statistical models only include robot actuation for modeling\cite{15CK}. Then, position feedback is involved in \cite{17AM} for adaptive modeling and control. Furthermore, temporal values are applied for control in \cite{16HW, 16WXb}. The authors of \cite{19GF} even include visual information for kinematic control. Recently, sensing information from various sensors like resistance and force sensors has been employed in \cite{23KT,20JL} for modeling adjustment. Most statistical approaches can be applied for both modeling and control with different inputs, like two SVR models in \cite{17AM}.}

\red{Besides data categories, the methods also evolve over time. For instance, the Jacobian matrix is involved in the EKF in \cite{11DR}, and the pseudo rigid robot model takes its place in \cite{18DL}. The authors of \cite{18ML} apply the adaptative Kalman filter, which shows strong robustness against the model nonlinearity and uncertainty instead of EKF. Recently, the unscented Kalman filter is leveraged in \cite{23KT}, which can apply the implicit Gaussian process for robot modeling. Also, GPR also evolves over time. The whole working space is divided into several regions, and each part requires a single GPR model for local modeling in \cite{19GF}. Then, only one online learning GPR model is employed to model the whole working space in \cite{20ZT}, and a meta-learning GPR model is employed in \cite{23ZT} for multiple new unknown working spaces.}

Statistical \red{methods} make \red{data distribution} assumptions from the perspective of statistics. They can attain an acceptable performance even with a small amount of data and become more effective with more data. Moreover, most of them can be leveraged for both modeling and control.
A comparison of some typical papers applying statistical \red{methods} is shown in Table \ref{table_s}. \red{This table first includes a simple regression model SVR in \cite{17AM}, then includes two regression approaches GPR in \cite{20ZT} and LWPR in \cite{17KL}. Finally, the observer EKF in \cite{16AA} is introduced.}
\input{tables/table_s}

\section{Neural Network}
\label{sec5}
\begin{figure*}[!ht]
\centering
\includegraphics[width=6.2in]{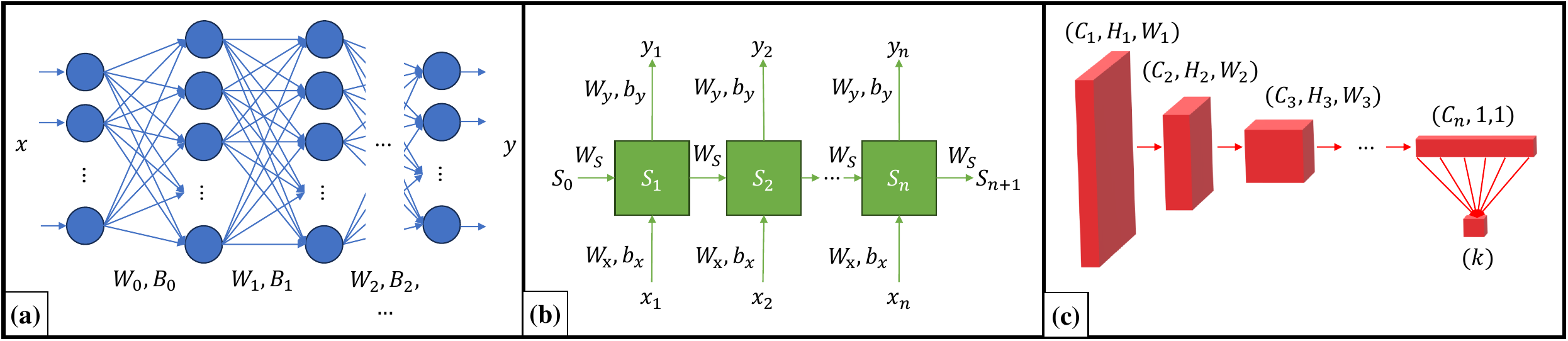}
\caption{Diagrams of (a) MLP, (b) RNN, and (c) CNN. MLP is composed of multiple layers. Parameters $W_*, B_*$ in one layer are in parallel and can be trained simultaneously. RNN takes the input data $x$ sequentially, and each bar shares the same weights $W_S, W_x, W_y, b_x,$ and $b_y$. \zc{CNN obtains matrix with channel $C_1$, height $H_1$, and width $W_1$ as input. In the network feedforward, the channel number improves while the width and height decrease using kernels. Finally, CNN outputs a matrix with dimension $(C_n,1,1)$, and a fully connected layer is employed to map to the target dimension $k$.}}
\label{fig6}
\end{figure*}

Considerable efforts have been focused on NN applications in the soft robot field. In the early years, extreme learning machines (ELM) \cite{13KN} and radial basis function (RBF) \cite{14AM} were popular choices. Nowadays, researchers prefer multilayer perceptron (MLP) \cite{13MG} and recurrent neural network (RNN) \cite{17TT} due to their generalized and sequence-related structures, respectively. Moreover, for some special proposes like image processing, autoencoder (AE) \cite{18GS} and convolutional neural network (CNN) \cite{19MA} are utilized. \red{Some typical NNs are shown in Fig. \ref{fig6}.}

\subsection{ELM and RBF}
\label{sec5.1}
ELM only comprises an input layer and an output layer. The input layer weights and bias are randomly assigned before training and fixed during training, while the output layer weights are trained to decrease loss. A simple loss is utilized in \cite{16WX} \red{for kinematics control}, which only aims to decrease the estimation error.  \red{The ELM is\begin{equation}
\label{eq5_1}
\begin{split}
\hat{a}=W^{out}\cdot f(W^{inp}\cdot p+B),
\end{split}
\end{equation}}
where $\hat{a}, p$ denote the estimated actuation value and the robot state, $W^{out},W^{inp}$ denote the output and input layer weights, $B$ denotes bias, and $f(x)={(1+e^{-x})}^{-1}$ is the sigmoid activation function. The training process of the original ELM in \cite{16WX} can be shown as
\begin{equation}
\label{eq5_2}
\begin{split}
\min_{W^{out}}\ & {\Vert A-\hat{A}\Vert},\\
s.t.\ &\hat{A}=W^{out}\cdot P,\\
\end{split}
\end{equation}
where $A$ and $P$ are all the real actions and input layer output in the training dataset, and $\hat{A}$ is the ELM estimation based on $P$. The optimized output layer weights are $\hat{W}^{out} = A \cdot P^ \dag$, where $P^ \dag$ is the pseudo‐inverse of the input layer output $P$.

In addition to estimation error, the authors of \cite{16RR,17RR} involve the norm of the output weight matrix into optimization error as a regularization term to avoid too large output weights \red{for kinematics control}. Constraints on the range of outputs are applied in \red{the ELM controller\cite{14JQ} according to the constraints of the real actuation. 
The training process can be shown as 
\begin{equation}
\label{eq5_3}
\begin{split}
\min_{W^{out}}\ & \sum^N_{i=0}{\Vert a_i-\hat{a}_i\Vert}^2+\alpha\cdot{\Vert W^{out}\Vert}^2,\\
s.t.\ &\hat{a}=W^{out}\cdot f(W^{inp}\cdot \mathbf{p}+B),\\
&\frac{\partial}{\partial p_i}\hat{a}(\mathbf{p})>0:\forall\mathbf{p}\in\Omega, i=1,\dots,n\\
&a_{min}<\hat{a}(\mathbf{p})<a_{max}:\forall\mathbf{p}\in\Omega, i=1,\dots,n\\
\end{split}
\end{equation}
where $a$ is the real actuation value, $\alpha$ is the regulation parameter, $N$ is the size of the training dataset, $n$ is the degrees of actuation, and $\Omega$ represents a set of samples basically cover the input space. The second condition guarantees that ELM has the same actuation direction as the real conditions, and the third is the actuation range constraint.} Recently, ELM has been utilized for online pose estimation in \cite{23XW} thanks to its simple structure and fewer parameters compared with other complicated NNs.

Based on the structure of ELM, RBF changes the activation function from a sigmoid activation function to multiple Gaussian functions \red{in \cite{14AM} for forward kinematic
modeling.} \red{Clustering can be applied to the training dataset to select the center of the Gaussian functions \cite{17RR}.} Although ELM and RBF include some basic elements of NN, e.g., activation functions and neurons, they contain some designed constraints for specific tasks and waste a part of their potential mainly due to the fixed parameters.

\subsection{MLP}
\label{sec5.2}
\red{MLP is the most popular NN. The diagram of MLP is shown in Fig. \ref{fig6}-(a). Generally, MLP can be shown as
\begin{equation}
\label{eq5_4}
\begin{split}
&y=\\
&W_n\cdot f(W_{n-1}\cdot f(\dots W_0\cdot f(x)+B_0\dots)+B_{n-1})+B_n,
\end{split}
\end{equation}
where $W_n, B_n$ represent network weight and bias of $n$-th layer, and $y,x$ are the output and input of MLP. To model a robot, the input is actuation, and the output is the robot state. For control, the desired and real robot states are input, and MLP can provide the estimated actuation.}

The first paper utilizing MLP in the soft robot field is \cite{07DB}, which is also the first paper utilizing NN. This work designs a particular parameter updating strategy \red{for control} instead of the backpropagation widely applied now. Similarly, training data is normalized in \cite{13MG} with principal component analysis (PCA) instead of batch and layer normalization, which are widely employed currently. MLP has many hyperparameters and changeable components, and plenty of papers have been conducted on them. For example, the authors of \cite{15MG} compare the performance of MLPs composed of different neuron numbers \red{on position control tasks}. Three activation functions (i.e., log-sigmoid, linear, and tan-sigmoid) are tested in \cite{18MW} for pneumatic robot modeling. The stochastic gradient descent (SGD) optimizer is applied in \cite{23TB} instead of the popular Adam optimizer. The authors of \cite{22XHb} comprehensively investigate the influence of hidden layer number, neuron number, and batch size on training time and validation loss \red{of sensing tasks}. 
In general, there is no optimal combination of the hyperparameters and component structures for all tasks, but most choices can obtain acceptable performance on most tasks. The detailed influence of each hyperparameter on the MLP performance has yet to be fully explored and explained.

Some uncommon MLPs have been applied to soft robots. The authors of \cite{19CC} provide targets to the hidden neurons of the MLP directly, similar to the conditional generative adversarial network (CGAN), which takes advantage of the information that is indirectly related to the mapping but restricts the potential of NN. A U-Net-like MLP is leveraged in \cite{20PH} for robot modeling in model predictive control (MPC), \red{ which connects the former and latter layers sequentially.} Multiple MLPs are connected in \cite{22GF}, which have different applications, e.g., forward kinematics modeling in simulation and sim-to-real transfer learning. \red{The combined MLP can collect actions in simulation and estimate the corresponding robot states. }

For soft robot modeling and control, some unique features of MLP are developed. 
For example, the inputs of some MLPs take physical models into consideration.
Based on the concentric tube model, the translations and rotations of the concentric tubes are employed into MLP inputs in \cite{20AK} \red{for modeling}. 
Furthermore, the authors of \cite{19RG} compare different joint space forms of the concentric tubes as input \red{on the forward modeling estimation tasks}.
Similarly, PCC is included in \cite{17HJ} by taking curvatures and curve lengths as MLP input.
\zc{The authors of \cite{23JL} utilize MLP to estimate physical parameters like mass inertia matrix, which is a model-based approach with NN.} 
Due to the time-delayed motion, end positions in the current and past timesteps are fed into the MLP in \cite{22FP} for control.

\subsection{RNN}
\label{sec5.3}
\red{RNN is a kind of NN that is designed specifically for sequential data. The diagram of RNN is shown in Fig. \ref{fig6}-(b). Although MLP can also receive temporal data, it is fed into networks simultaneously and fails to infer the sequential relationship. RNN takes data in sequence using recurrent structure. The $n$-th bar in Fig. \ref{fig6}-(b) can be represented as
\begin{equation}
\label{eq5_5}
\begin{split}
S_n&=f(W_x\cdot x_{n}+W_s\cdot S_{n-1}+b_{x}), \\
y_n&=g(W_y\cdot S_n+b_{y}),\\
\end{split}
\end{equation}
where $W_*, b_*$ are the RNN weight and bias parameters, $S_n$ is the hidden state of the step $n$, and $f(),g()$ are activation functions. As shown in Fig. \ref{fig6}-(b), the network of each step takes the hidden state from the previous step. Such a structure infers the sequential relationship of data while the other NNs take the data input simultaneously. In this case, RNN is more suitable than the other networks like MLP and CNN for soft robots, which provide delayed responses. The modeling and control with RNN share the same data category requirement with MLP mentioned in Eq. \ref{eq5_4}, but RNN needs data from multiple time steps instead of a single time step.}

The first RNN applied is modified Elman NN in \cite{15AM}, which restores information in previous steps with context nodes \red{for dynamic control}. Then, researchers leverage a nonlinear autoregressive network with exogenous inputs (NARX) in a series of papers\cite{17TT,17TTb,18TT} \red{for dynamic control}. This kind of RNN receives outputs in previous steps as a part of input in the current step.

Long short-term memory (LSTM) is a kind of RNN proposed for long-term dependence issues. This network has been used on endoscopic robot distal force prediction \cite{19XL,23SY} and external force position and magnitude prediction \cite{21ZD}. Moreover, such a network can cope with sensing signals from nonlinear time-variant soft sensors and achieves tasks like position prediction \cite{19TTb}, object recognition\cite{22ZZ}, and shape reconstruction \cite{18GS}. In addition to perception, the authors of \cite{22DW} \red{dynamically} control a robotic catheter with LSTM to decrease contact force. \red{The authors of \cite{23ZCc} employ LSTM as an offline dynamic controller to cope with the nonlinear behaviors.} In summary, RNNs perform satisfactorily on various tasks due to their sequential structures and memory ability.

\subsection{Special NN}
\label{sec5.4}
There are some NNs that take advantage of visual information in soft robot control. For example, AE is utilized to extract features from the images of soft robots in \cite{18GS} and estimate the robot's shape. \zc{CNN estimates shapes\cite{23EAb} and joint values\cite{21NL} based on robot images.} The diagram of CNN is shown in Fig. \ref{fig6}-(c). Also, CNN can \red{predict the orientation} of the placenta in \cite{19MA} and encodes robot deformation for \red{shape estimation} with the help of markers inside the chamber in \cite{23UY}. Furthermore, although RNN is the most popular choice for sequential information processing, CNN can also be applied \red{for dynamic modeling} using rearranged pressure inputs \cite{19YZ}. \red{The space relationship in a matrix is leveraged to infer the sequential relationship of an actuation sequence for hysteresis modeling.} A 3D NN is employed in \cite{17YC} for a segmental surgical manipulator, which considers the time sequence between layers and the segmental sequence between neurons \red{for planning.} \zc{Spatial sequences of soft modular robots are considered in \cite{23ZCd} by utilizing bidirectional RNN. A generative adversarial network (GAN) is utilized for synthetic data in \cite{23SS}.}

\subsection{Summary Neural Network}
\label{sec5.5}
\red{The application of NN in soft robots started with \cite{07DB}. First, some simple networks like ELM and RBF are included. Then, with the development of NN, the researchers give up the constraints of ELM, change the activation function of RBF, and enlarge the network. In this case, MLP becomes a good tool for both modeling and control. Furthermore, due to the hysteresis of soft robots, RNN is applied to deal with time-related data. AE and CNN are employed to process images.}

To summarize, owing to the large variety of structures, NN is attractive to soft robot research. For most issues in soft robot modeling and control, it is highly possible to find a related NN solution. However, such models require a large amount of data due to their complicated structures, and it is challenging to update them online. A comparison of some typical papers applying NNs is shown in Table \ref{table_n}. \red{This table begins with a simple NN, RBF\cite{14AMb}. Then, a common NN, MLP\cite{22XHb}, is included in this table. Finally, LSTM for control\cite{22DW} and CNN for modeling\cite{19YZ} are summarized in Table \ref{table_n}.}
\input{tables/table_n}

\section{Reinforcement Learning}
\label{sec6}
RL copes with high-level tasks by exploring the environment and exploiting data collected during exploration. This strategy trains an agent for complicated tasks and requires a long learning time \red{and a massive amount of data. The agent is trained considering defined reward functions. This approach cannot be used for modeling. Moreover, it requires exploring environments, which may be provided by modeling methods.}

\subsection{RL implementation}
\label{sec6.1}
\red{In the early years, statistical models were applied as the agents in RL instead of NN.} A GPR model named Gaussian process temporal difference method is employed in \cite{05YE} to control an octopus arm. As the RL agent, a GMM is trained to estimate robot shape and contact in \cite{13YK} and control a flexible surgical robot to go through a tube in \cite{16JC}. For robotic catheter control inside a narrow tube, a joint probability distribution is learned considering various variables like tip and entrance points, touch state, and action in \cite{14AT}. Q-learning is implemented with a Q table or Q function as the agent. 
\red{The agent decides the action $A$ based on the state $S$ and a certain policy like $\epsilon$ greedy. The training process of the Q table or function can be shown as
\begin{equation}
\label{eq6_1}
\begin{split}
&Q(S_t, A_t) \leftarrow \\
&\alpha[R(S_t, A_t)+\gamma \max_{a}{Q(S_{t+1}, a)}-Q(S_t, A_t)] + Q(S_t, A_t),
\end{split}
\end{equation}
where $S_t$ is the state at time $t$, $A_t$ is the action decided by the Q table or function according to the state $S_t$, and $R(S_t, A_t)$ is the reward function. $\alpha$ and $\gamma$ are the learning rate and discount rate, respectively. $\max \limits_{a}{Q(S_{t+1}, a)}$ means the maximal Q value for the state $S_{t+1}$ and every possible action $a$.} For example, the authors of \cite{21HJ} exploit Q-learning for many sophisticated control tasks like turning a handwheel, unscrewing a bottle cap, drawing a line with a ruler, and so on.

With the help of NN as an agent, RL not only achieves simple tasks like position reach \cite{22XD} and trajectory following \cite{21GJ} but also addresses some complex issues like gait design \cite{20XL}. A soft snake robot is controlled to move on the ground and arrive at target positions in \cite{20XL}. It is challenging to control the robot gait with traditional control methods, but RL is utilized for gait design and obstacle avoidance in \cite{21XL}. The authors of \cite{23YL} fuse the visual and shape information with NN in RL and control a flexible endoscopy to navigate.

\subsection{RL in soft robots}
\label{sec6.2}
\red{Soft robots can take advantage of various unique RL strategies specifically for soft robots to cope with some challenges.} The infinite DOFs lead to wide actuation and state spaces, which bring a burden to explore the whole environment. In this case, most researchers discretize these spaces. The action space in \cite{05YE} is restricted to only six available actions, and the authors of \cite{19SS} discretize the workspace into a 3D grid with a resolution of 0.01 m. Although discretization limits the RL potential, it produces simple spaces and decreases the training time. The soft robot in \cite{22YG} is able to keep the end position invariant while changing the orientation with the help of RL.

One of the most considerable challenges of RL is exploring the real world, which has a high time cost and may damage robots. Therefore, modeling in simulation, especially with physical methods, is widely utilized in the training process. \red{The authors of \cite{22OO}} first train an RL strategy to control a slave robot in a simulator named CoppeliaSim and then test it on the real robot. Constant curvature (CC), a soft robot modeling method, provides a simulation environment for \cite{22YL} at first, and the NN agent continues to learn in the real world using the Deep Deterministic Policy Gradient method (DDPG). In most cases, the explored environment is modeled by a trained NN. For example, MLP is applied for modeling in \cite{20QW} as the exploring environment. Also, RNNs are utilized in \cite{19TT,22AC}, which are NARX and LSTM respectively, for forward modeling of segmented pneumatic robots. Then, RL agents are trained and validated in reality.

\subsection{Summary Reinforcement Learning}
\label{sec6.3}
\red{RL application in soft robots first starts with the help of statistical models as agents, and then NNs take the place due to their generalization. Discretization is widely applied for RL in soft robots. Meanwhile, RL leverages soft robot simulators and stimulates their development. With such RL strategies, now soft robots can achieve some complicated tasks.}

Compared with other approaches, RL requires the most enormous amount of data. More critically, a predefined agent and interaction with the environment are necessary. Following such a high cost, RL can fulfill complex and high-level tasks. With some adaptation strategies like discretization and simulation transfer learning, the time and resource costs can be reduced to some extent. A comparison of typical papers applying RL is shown in Table \ref{table_r}. \red{This table begins with an RL strategy applying the statistical model Gaussian process as agent\cite{05YE}, and the authors of \cite{21GJ} apply NNs for RL. Then, two strategies for soft robot RL are included. RL strategies like discretization are employed in \cite{19SS}, and the authors of \cite{22YL} pretrain the agent in simulation and test it in reality.}

\input{tables/table_r}
\input{tables/table_19}

\section{Conclusions and Discussions}
\label{sec7}
\red{In this section, we summarize the foundations, data-driven methods, and their representative papers in Subsection \ref{sec7.1}. The benefits and limitations of data-driven approaches involved in the review are included and compared in Subsection \ref{sec7.2}. Finally, we forecast the emerging directions for soft robot research in Subsection \ref{sec7.3}.}

\subsection{\red{Summary}}
\label{sec7.1}
\red{This review summarizes the data-driven approaches applied in soft robot modeling and control.
The physical approaches provide simulators for data-driven methods, like SOFA\cite{12FF}, Pyelastica\cite{18MG}, and SoRoSim\cite{22AM}. 
The Jacobian \zc{matrices} describe soft robots and control the robots with the inverse Jacobian matrix or optimization. The authors of \cite{14MY} firstly utilize the Jacobian matrix in the main approach.}

Statistical models aim to achieve modeling and control with datasets. Regression methods like GPR\cite{19GF} and LWPR\cite{17KL} can estimate the mapping functions, and each trained model can be applied for either modeling or control according to the input of the training data. GMM\cite{19BY} encodes the dataset into a joint data distribution, which can be applied for both modeling and control. Observers like EKF\cite{23KT} can estimate robot states based on one existing modeling method.
Recently, NN has been the most compelling tool for soft robots. ELM\cite{14JQ} was a popular choice in early years, and now MLP\cite{15MG}, which is more complicated, and RNN\cite{18TT}, which can cope with time-related data, have shown their benefits. Similar to statistical models, NN can also be applied for both modeling and control. RL shows good performance on position control\cite{22AC}, planning\cite{23KI}, and even some sophisticated manipulation tasks\cite{21HJ}. Generally, RL does not provide a robot model but exploits an existing environment. The papers based on data-driven methods, which are cited in this review and published after 2019, are summarized in Table \ref{table_19}.

\red{It is apparent that sophisticated models like RL require a larger amount of data while achieving better performance, but they also improve the computation cost.
Oversimplified approaches like the Jacobian matrix are only feasible for limited simple tasks, but they are easy to understand and can achieve a high control frequency. 
Considering both cost and performance, each model has its own advantages and disadvantages, and there is no optimal solution for all tasks.}

\subsection{\red{Advantages and Limitations}}
\label{sec7.2}
\red{Statistical modeling and control approaches are moderate and flexible approaches in these methods. Generally, they only require a moderate amount of data, which is less than the NN requirement but more than the requirement of the Jacobian matrix. Also, the control frequency is lower than that of the Jacobian matrix but higher than the RL frequency. One statistical model can be applied for both modeling and control online, but some models are only local models and lack generalization ability, which can be proven by the working space segmentation in \cite{19GF}.}
\red{NN is very suitable for soft robot modeling and control due to the nonlinear activation functions and complex network structure. This model shows its general applicability. However, the data amount requirement accelerates the aging of soft robots and takes a long time for data collection. NN cannot be a fast and online approach due to the slow training process.
By exploring the environment and exploiting the data, RL can achieve some high-level tasks like navigation. Careful planning is not required even for sophisticated tasks. Meanwhile, RL has a high requirement for time and computation resources. Exploring the real world may damage the robots, and some environments, like the human body, cannot be applied for exploring. Highly realistic simulators are required if the RL agent is trained in simulation. To summarize, RL is a useful but consuming approach.}

\subsection{\red{Emerging Directions}}
\label{sec7.3}
\red{In the other research areas like computer vision and natural language processing, the original NN is simple at first\cite{98YL}. Then, more complex NNs with large sizes and different structures are proposed, like ChatGPT, BERT\cite{18JD} and YOLO\cite{16JR}. Similarly, the size and complexity of NN in soft robots will improve. Such models can achieve more challenging tasks compared with simple NNs.
However, large models lead to low frequency for control implementation, and one should consider balancing the model complexity and computation cost based on the modeling and control requirement.}

\red{The research on soft robots begins with the application of one single approach and simple tasks.
Recently, there have been some papers combining multiple methods to solve difficult problems like model mismatch. For example, the authors of \cite{23ZCc} apply offline RNN and online kinematics model for control. MPC and iterative learning controller are combined in \cite{19ZT}. Two NNs are included in \cite{23MN} for RL agent and model mismatch adjustment. 
As discussed in Subsection \ref{sec7.2}, each method shows its own advantages and disadvantages. The usage of multiple methods can take advantage of every approach and achieve better performance.}

The medical environment, as one of the most significant applications of soft robots, has a high standard for safety, efficiency, and convenience. To involve soft robots in medical applications, some advanced modeling and control strategies are required. Although so many works achieve controlling the robot end pose, it is challenging to control the whole robot shape and avoid contact which may damage the human body.
NN and RL lack interpretability and are challenging to apply in real surgery. Also, it is impossible for RL to explore \textit{in vivo}, and RL agents can only be trained in simulation or physical simulators. The automatic medical soft robot control is still in its nascence from the aspect of safety and data requirements.
Cooperation among robotics researchers, doctors, and related departments is required to address these issues.

\bibliographystyle{IEEEtran}
\bibliography{IEEEabrv,references}
\vfill

\begin{IEEEbiography}[{\includegraphics[width=1in,height=1.25in,clip,keepaspectratio]{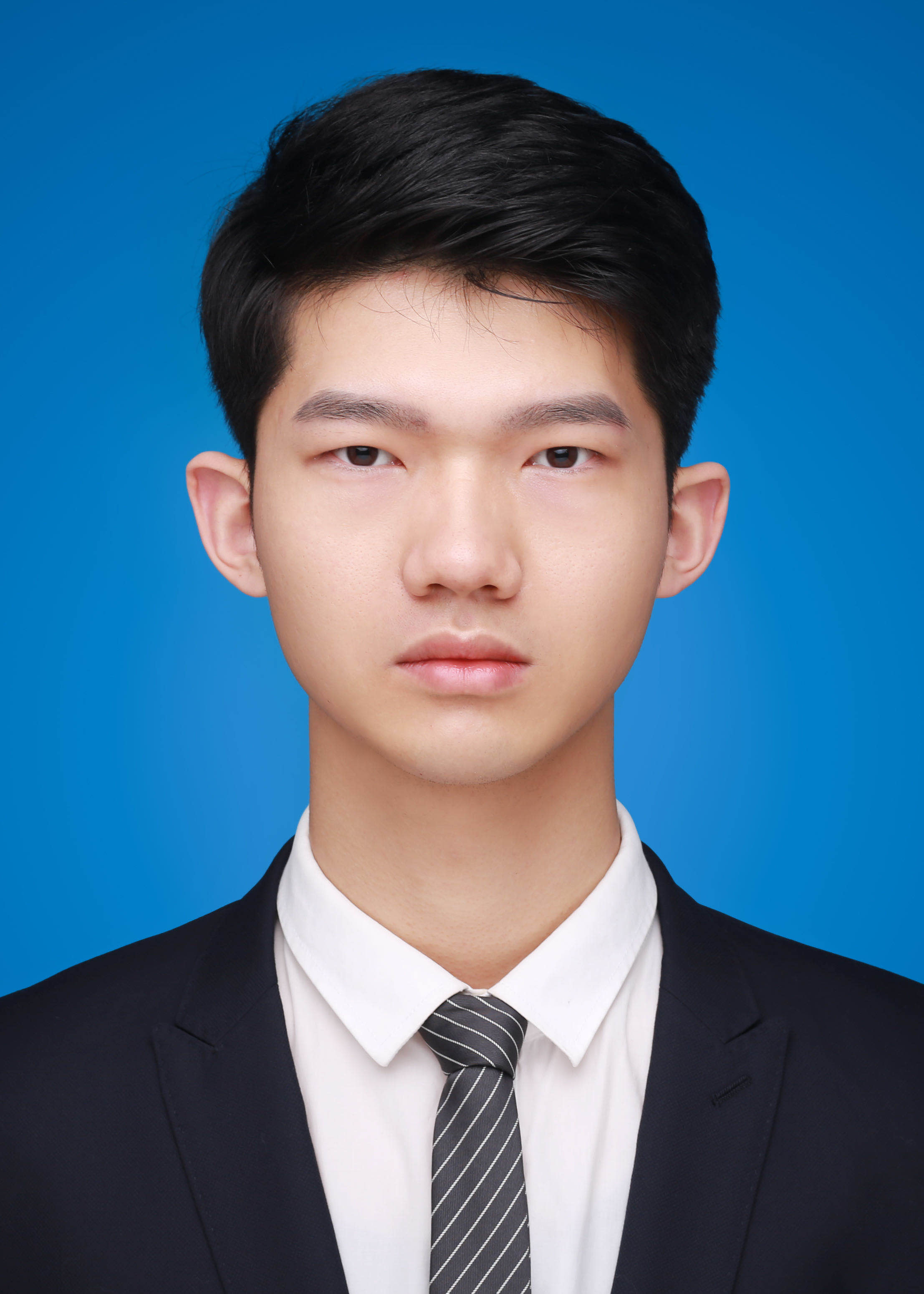}}]{Zixi Chen} received the M.Sc. degree in Control Systems from Imperial College in 2021. He is currently pursuing the Ph.D. degree in Biorobotics from Scuola Superiore Sant’Anna of Pisa.

His research interest includes optical tactile sensors and soft robot control with neural networks.
\end{IEEEbiography}

\begin{IEEEbiography}[{\includegraphics[width=1in,height=1.25in,clip,keepaspectratio]{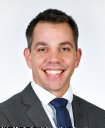}}]{Federico Renda} (Member, IEEE) received the B.Sc. and M.Sc. degrees in biomedical engineering from the University of Pisa, Pisa, Italy, in 2007 and 2009, respectively, and the Ph.D. degree in biorobotics from the Biorobotics Institute, Scuola Superiore Sant’Anna, Pisa, in 2014.

He is currently an Associate Professor with the Department of Mechanical Engineering, Khalifa University, Abu Dhabi, UAE. His research interests include dynamic modeling and control of soft and continuum robots using principles of geometric mechanics.
\end{IEEEbiography}

\begin{IEEEbiography}[{\includegraphics[width=1in,height=1.25in,clip,keepaspectratio]{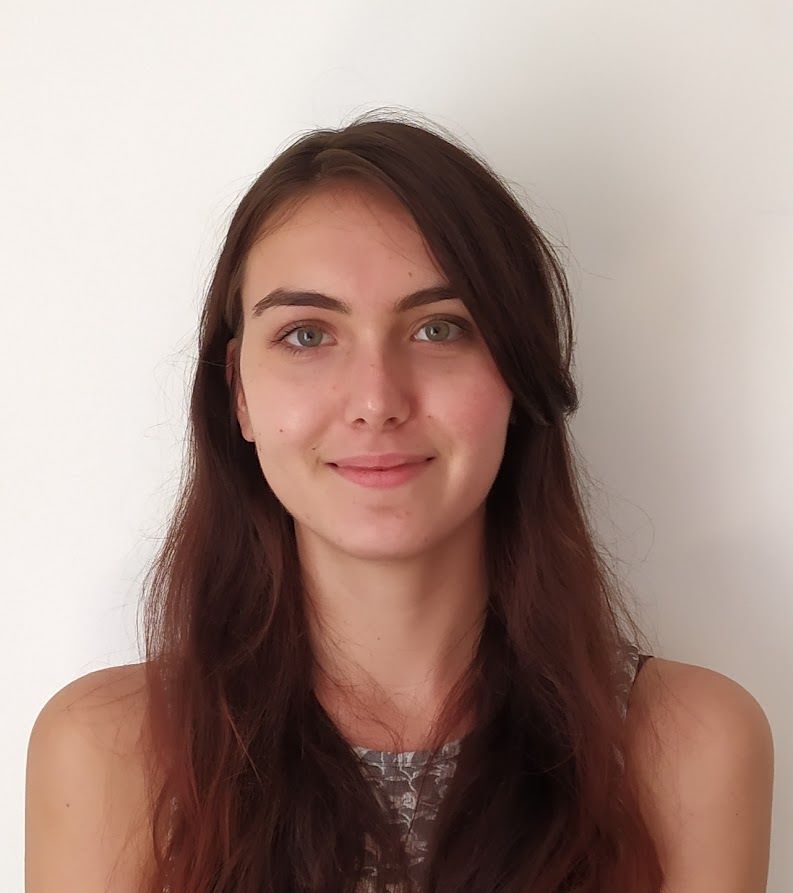}}]{Alexia Le Gall} received her Engineering degree in Mechanical Conception and M.Sc. degree in Mechatronics from the University of Technology of Compiègne (France) in 2022. She is currently pursuing a Ph.D. in BioRobotic from the Scuola Sant'Anna of Pisa, with a specialisation in Soft Robotic.

Her interest includes mechanical design and soft robotic actuation and fabrication methods.
\end{IEEEbiography}

\begin{IEEEbiography}
[{\includegraphics[width=1in,height=1.25in,clip,keepaspectratio]{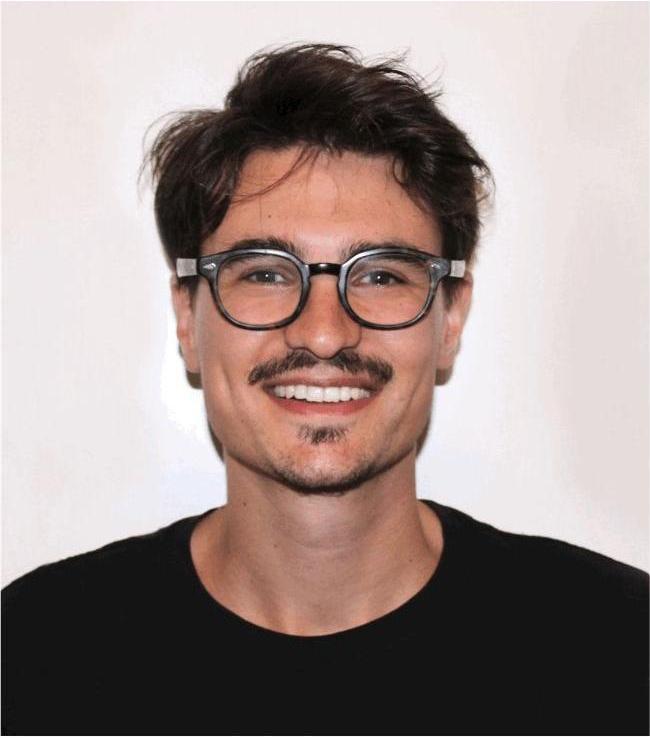}}]{Lorenzo Mocellin} is a Ph.D. candidate at The BioRobotics Institute, Scuola Superiore Sant'Anna. He currently works in the Surgical Robotics Lab, on design and development of surgical robotic tools.

He received the B.Sc. degree in Biomedical Engineering from University of Padua in 2019 and the M.Sc. in Bioengineering from University of Trieste in 2022, with a thesis on cardiovascular fluid mechanics modeling. 
His research focuses on miniaturized medical devices, surgical robotic end-effectors, origami-inspired structures, compliant mechanisms, and biomaterials.
\end{IEEEbiography}

\begin{IEEEbiography}[{\includegraphics[width=1in,height=1.25in,clip,keepaspectratio]{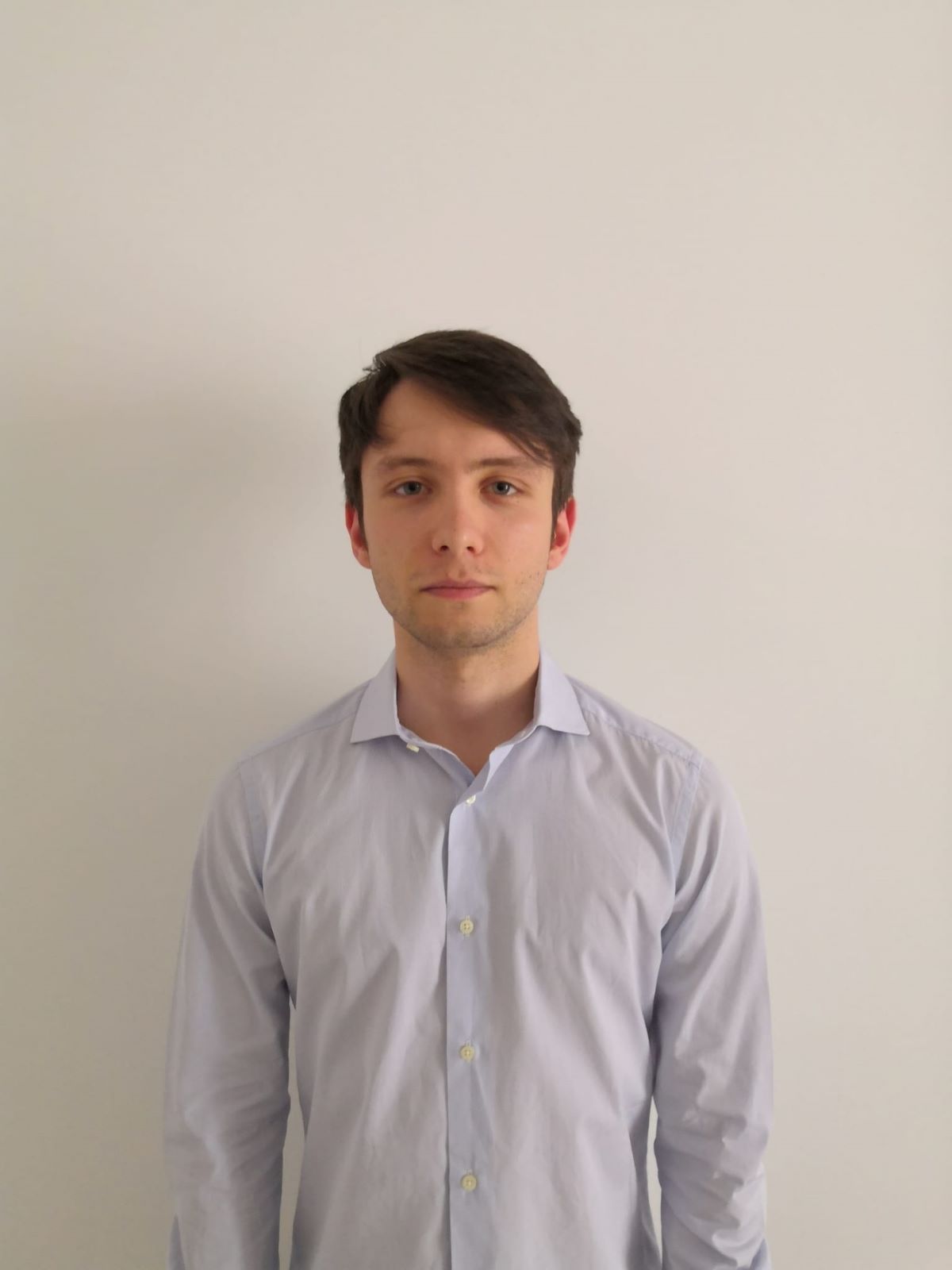}}]{Matteo Bernabei} is a Ph.D. student in Biorobotics at Scuola Superiore Sant'Anna (SSSA), Italy, since 2023. He received a Master's degree in Biomedical Engineering in 2022 and a Bachelor's degree in Industrial Engineering in 2020, both at Campus Bio-Medico University in Rome, Italy. In 2021, as part of his formation, he spent 6 months as research intern at Boston Children's Hospital to work on advanced methods for analysis of EEG signals for the optimization of pre-surgery planning in children with drug resistant epilepsy. His interests now include endoluminal and minimally invasive surgical robotics.
\end{IEEEbiography}

\begin{IEEEbiography}[{\includegraphics[width=1in,height=1.25in,clip,keepaspectratio]{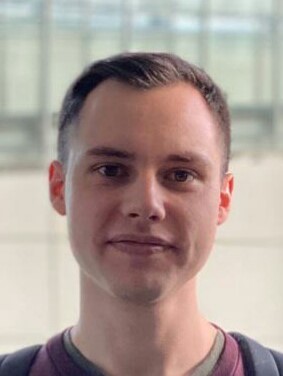}}]{Théo Dangel} received an M.Eng. in mechanical engineering from SupMicroTech-ENSMM in Besançon, France and an M.S in electronics from Tokyo Denki University in Tokyo, Japan, both in 2022. He is currently pursuing a Ph.D. degree in biorobotics at Scuola Superiore Sant’Anna in Pisa, Italy.
His research interest includes actuators for soft robotic.
\end{IEEEbiography}

\begin{IEEEbiography}[{\includegraphics[width=1in,height=1.25in,clip,keepaspectratio]{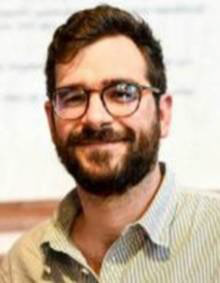}}]{Gastone Ciuti} (Senior Member, IEEE) received the master’s degree (Hons.) in biomedical engineering from the University of Pisa, Pisa, Italy, in 2008, and the Ph.D. degree (Hons.) in biorobotics from The BioRobotics Institute of Scuola Superiore Sant’Anna, Pisa, Italy, in 2011. He is currently an Associate Professor of Bioengineering at Scuola Superiore Sant’Anna, leading the Healthcare Mechatronics Laboratory. He has been a Visiting Professor at the Sorbonne University, Paris, France, and Beijing Institute of Technology, Beijing, China, and a Visiting Student at the Vanderbilt University, Nashville, TN, USA, and Imperial College London, London, U.K. He is the coauthor of more than 110 international peer reviewed papers on medical robotics and the inventor of more than 15 patents. His research interests include robot/computer-assisted platforms, such as tele-operated and autonomous magnetic-based robotic platforms for navigation, localization and tracking of smart and innovative devices in guided and targeted minimally invasive surgical and diagnostic applications, e.g. advanced capsule endoscopy. He is a Senior Member of the Institute of Electrical and Electronics Engineers (IEEE) society and Member of the Technical Committee in BioRobotics of the IEEE Engineering in Medicine and Biology Society (EMBS). He is an Associate Editor of the IEEE Journal of Bioengineering and Health Informatics, IEEE Transaction on Biomedical Engineering and of the IEEE Transaction on Medical Robotics and Bionics.
\end{IEEEbiography}

\begin{IEEEbiography}[{\includegraphics[width=1in,height=1.25in, clip,keepaspectratio]{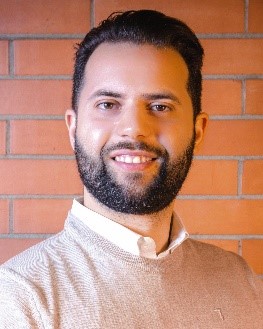}}]{Matteo Cianchetti} is Assistant Professor at The BioRobotics Institute of Scuola Superiore Sant’Anna, leading the Soft Mechatronics for Biorobotics Lab. He received the MSc degree in Biomedical Engineering (cum laude) from the University of Pisa, Italy, in 2007 and the PhD degree in Biorobotics from Scuola Superiore Sant’Anna. He is author or co-author of more than 100 international peer reviewed papers and he regularly serves as a reviewer for more than 10 international journals. He is Chair of the IEEE TC on Soft Robotics and he has been and currently is involved in EU-funded projects with the main focus on the development of Soft Robotics technologies. His main research interests include bioinspired robotics and the study and development of new systems and technologies based on soft/flexible materials for soft actuators, smart compliant sensors and flexible mechanisms.
\end{IEEEbiography}

\begin{IEEEbiography}[{\includegraphics[width=1in,height=1.25in,clip,keepaspectratio]{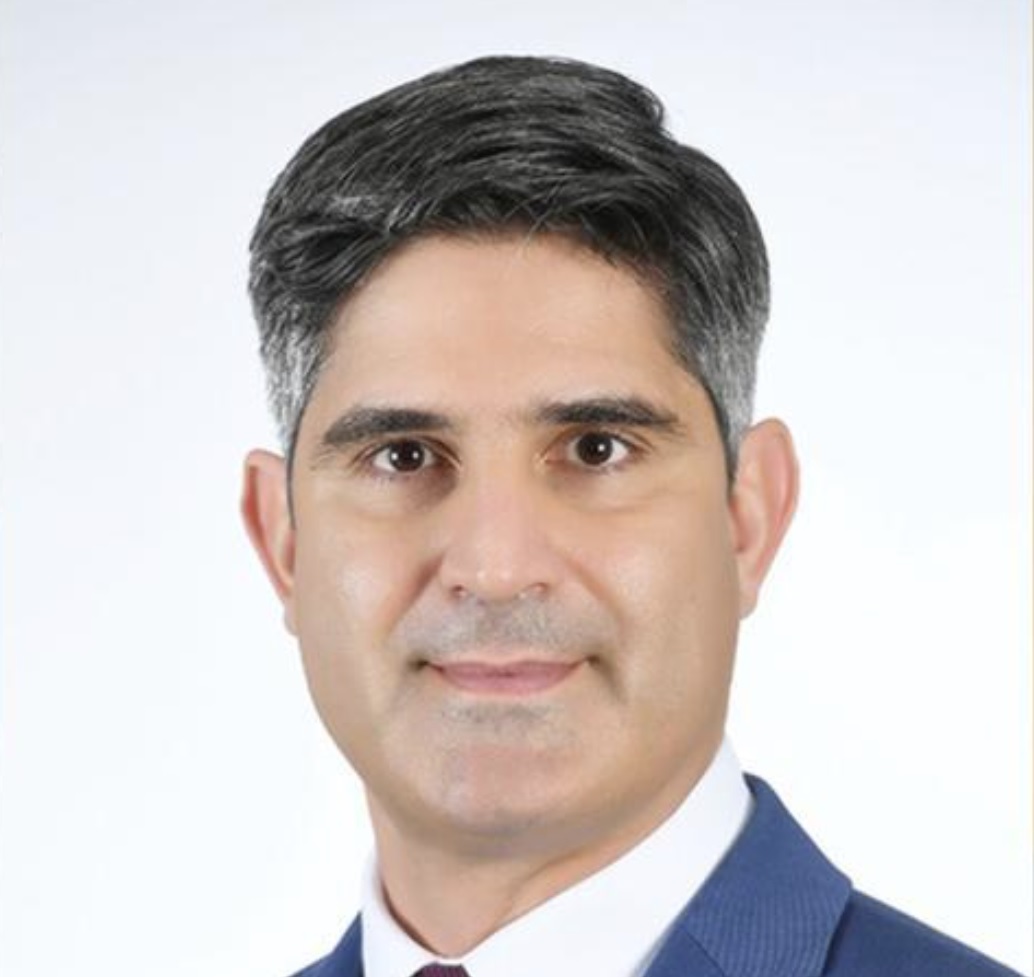}}]{Cesare Stefanini} (Member, IEEE) received the M.Sc. degree in mechanical engineering and the Ph.D. degree in microengineering, both with honors, from Scuola Superiore Sant-Anna (SSSA), Pisa, Italy, in 1997 and 2002, respectively.

He is currently Professor and Director of the BioRobotics Institute in the same University where he is also the Head of the Creative Engineering Lab. His research activity is applied to different fields, including underwater robotics, bioinspired systems, biomechatronics, and micromechatronics for medical and industrial applications. He received international recognitions for the development of novel actuators for microrobots and he has been visiting Researcher with the University of Stanford, Center for Design Research and the Director of the Healthcare Engineering Innovation Center, Khalifa University, Abu Dhabi, UAE.

Prof. Stefanini is the recipient of the “Intuitive Surgical Research Award.” He is the author or coauthor of more than 200 articles on refereed international journals and on international conferences proceedings. He is the inventor of 15 international patents, nine of which industrially exploited by international companies. He is a member of the Academy of Scientists of the UAE and of the IEEE Societies RAS (Robotics and Automation) and EMBS (Engineering in Medicine and Biology).
\end{IEEEbiography}
\end{document}

%% file: tables/table_sum.tex
\begin{table*}[!htb]
\caption{Summary of \red{Method} Categories}
\centering
\begin{tabular}{p{0.35in} p{0.7in} p{0.9in} p{1.8in} p{0.4in} p{0.4in} p{1.4in}}
\red{Method} & Data amount & Modeling & Controller & Online updating & Dynamic control & Control feature\\
\hline
Statistical model & A moderate amount & Forward models (GPR, GMM, etc.) & Most models (GPR, GMM, etc.) and optimization based on forward models & \Checkmark & \Checkmark & Extract features only from data, forward and inverse \red{mapping} with the same model\\
NN & A large amount & Forward models & Models similar to modeling networks & \XSolid & \Checkmark & Various structures and tasks\\
RL & A huge amount  & Explored environment & Agent & \XSolid & \Checkmark & Exploring environment and exploiting data, high level tasks\\

\end{tabular}
\label{table_sum}
\end{table*}

%% file: tables/table_p.tex
\begin{table}[!htb]
\caption{Physical Model Paper Comparison}
\centering
\begin{tabular}{p{0.5in} p{0.5in} p{0.5in} p{1.2in} p{1.2in}}
Publication & Simulator & Simulation approach & Supplementary introduction \\
\hline
Faure \emph{et al.}\cite{12FF} &  SOFA & FEM & Widely applied in organ simulation\\

Gazzola \emph{et al.}\cite{18MGb} & PyElastica & Cosserat rod & Validated with soft filaments \\

Mathew \emph{et al.}\cite{22AM} &  SoRoSim & GVS & Integrated into a user-friendly MATLAB toolbox\\

Iyengar \emph{et al.}\cite{23KI} & Customized & Concentric tube model & RL is applied for control
\end{tabular}
\label{table_p}
\end{table}

%% file: tables/table_j.tex
\begin{table*}[!htb]
\caption{Jacobian \zc{Matrix} Paper Comparison}
\centering
\begin{tabular}{p{0.55in} p{0.4in} p{0.6in} p{0.65in} p{0.8in} p{0.5in} p{1.0in} p{1.2in}}
Publication & Forward Model & Control method & Control target & Main controller &
Control frequency & Jacobian matrix updating input & Controller details  \\
\hline
Yip and Camarillo\cite{14MY} & Jacobian matrix & Optimization & Position control & Jacobian model&
1kHz & End effector positions and tension & Fast online controller\\

Yip and Camarillo\cite{16MY} & Jacobian matrix & Optimization & Force and position control & Jacobian model&
$\geq$2kHz & End effector forces, positions and tension & Controller for position and force\\

Jiang \emph{et al.}\cite{21HJ} & Jacobian matrix & Inverse Jacobian matrix & High-level tasks & RL &
2Hz & End effector pose and pressure & Jacobian model applied as the low-level controller

\end{tabular}
\label{table_j}
\end{table*}

%% file: tables/table_s.tex
\begin{table*}[!htb]
\caption{Statistical \red{Method} Paper Comparison}
\centering
\begin{tabular}{p{0.88in} p{0.4in} p{0.65in} p{0.45in} p{0.3in} p{0.35in} p{0.3in} p{2.3in}}
Publication & Modeling & Control & Control frequency & Data amount &
Dynamic control & Online update & Controller details  \\
\hline
Melingui \emph{et al.}\cite{17AM} & SVR & The other SVR model & - & 4096 & \XSolid & \Checkmark & Adaptative controller by online controller updating\\
Tang \emph{et al.}\cite{20ZT} & GPR & Optimization & 10Hz & 120 &
\Checkmark & \Checkmark & Initialization in simulation and learning in reality\\
Lee \emph{et al.}\cite{17KL} & LWPR & The same LWPR model & 20Hz & - &
\Checkmark & \Checkmark& Online controller to adapt distrubance\\
Ataka \emph{et al.}\cite{16AA} &  EKF & Inverse Jacobian matrix & 40Hz & - &
\Checkmark & \Checkmark& Controller for segmented robot and obstacle avoidance\\
\end{tabular}
\label{table_s}
\end{table*}

%% file: tables/table_n.tex
\begin{table*}[!htb]
\caption{Neural Network Paper Comparison}
\centering
\begin{tabular}{p{0.7in} p{0.4in} p{0.7in} p{0.3in} p{0.85in} p{0.4in} p{0.3in} p{0.35in} p{1.5in}}
Publication & Modeling & Control & Hidden layer & Activation function &
Neuron & Data amount & Dynamic control & Controller details  \\
\hline
Melingui \emph{et al.}\cite{14AMb} & RBF & The other RBF & 1 &  Gaussian function &
72/74 & 4096 & \XSolid & Forward and inverse model with the similar NN structure\\
Ha \emph{et al.}\cite{22XHb} & MLP &  N/A & 1-4 & ReLU &
15-720 & 28738 & \XSolid & Hyperparameter finetuning\\
Wu \emph{et al.}\cite{22DW} & N/A &  LSTM & 4 & Tanh/Sigmoid &
128 & 260269 & \Checkmark & Dynamic controller and minimizing contact force\\
Zhang \emph{et al.}\cite{19YZ} & CNN & N/A &4 & Tanh &
$2\times2\times4$ & 3800 & \Checkmark & CNN dynamic modeling with transformed data\\
\end{tabular}
\label{table_n}
\end{table*}

%% file: tables/table_r.tex
\begin{table*}[!htb]
\caption{Reinforcement Learning Paper Comparison}
\centering
\begin{tabular}{p{0.7in} p{0.4in} p{0.4in} p{0.35in} p{0.3in} p{0.35in} p{0.6in} p{0.6in} p{1.8in}}
Publication & Modeling & Control & Agent & Episode & Learning strategy & Training environment &
Testing environment & Controller details\\
\hline
Engel \emph{et al.}\cite{05YE} & FEM & RL & Gaussian process & 3000 &  GPTD & Sim & Sim & The first data model applied for soft robot control \\

Ji \emph{et al.}\cite{21GJ} & CC & RL & NN & 100 & MADQN & Sim and Real &
Sim and Real & Agent composed of evaluation and target network\\

Satheeshbabu \emph{et al.}\cite{19SS} & N/A & RL & NN & 5000 & DQN & Real &
Real & Discrete task space\\

Li \emph{et al.}\cite{22YL} & CC & RL & NN & 500 & DDPG & Sim &
Sim and Real & Pre-training in Sim and learning in Real\\
\end{tabular}
\label{table_r}
\end{table*}

%% file: tables/table_19.tex
\begin{table*}[!htb]
\caption{Summary of Papers after 2019}
\centering
\begin{tabular}{p{0.9in} p{1.2in} p{3.5in}}

Data model & Category & Publications  \\
\hline


Statistical \red{method} & SVR & \cite{20IL}\\
& kNN & \cite{22XH}\\
& KF & \cite{19MV,20JL,23KT}\\
& Gaussian Distribution & \cite{21ZD}\\
& GPR & \cite{20JJ, 20ZT, 22YL, 22ZT, 22ZTb, 23KT, 23XW, 23ZT}\\

NN & ELM & \cite{23XW}\\
& MLP & \cite{20AK, 20IL, 20JB, 20QW, 20XW, 21GJ, 21NL, 22AC, 22FP, 22GF, 22OO, 22XD, 22XHb, 22YL, 22ZD, 23JL, 23KI, 23TB, 23UY}\\
& RBF & \cite{20IL}\\
& LSTM & \cite{21ZD, 22DW, 22ZZ, 23SY, 23ZCc}\\
& Special NN & \cite{22XH} (AE), \cite{20PH} (U-Net), \cite{20XL,21XL} (CPG), \cite{21NL,23YL,23UY,23EAb} (CNN), \cite{23SS} (GAN), \cite{23ZCd} (biRNN)\\

RL & Q-learning & \cite{21HJ, 22YG}\\
& DQN & \cite{20QW,22OO}\\
& MADQN & \cite{21GJ}\\
& PPO & \cite{20XL,21XL}\\
& TRPO & \cite{22AC}\\
& DDPG & \cite{22XD,22YL}\\

\end{tabular}
\label{table_19}
\end{table*}